\documentclass[runningheads]{llncs}
\usepackage{graphicx}

\usepackage{tikz}
\usepackage{comment}
\usepackage{amsmath,amssymb} 
\usepackage{color}
\usepackage{caption}
\usepackage{subcaption}
\usepackage{multirow}

\usepackage[accsupp]{axessibility}  

\usepackage[pagebackref,breaklinks,colorlinks]{hyperref}

\begin{document}
\pagestyle{headings}
\mainmatter
\def\ECCVSubNumber{7283}  

\title{3DG-STFM: 3D Geometric Guided Student-Teacher Feature Matching} 

\titlerunning{3DG-STFM}
%
\author{Runyu Mao\inst{1}\and
Chen Bai\inst{2}\and
Yatong An\inst{2}\and
Fengqing Zhu\inst{1}\ and
Cheng Lu\inst{2}}
\authorrunning{R. Mao et al.}
%
\institute{Purdue University
\email{\{mao111, zhu0\}@purdue.edu}\and
XPeng Motors
\email{\{chenbai, yatongan, luc\}@xiaopeng.com}
}
\maketitle

\begin{abstract}
We tackle the essential task of finding dense visual correspondences between a pair of images.
This is a challenging problem due to various factors such as poor texture, repetitive patterns, illumination variation, and motion blur in practical scenarios.
In contrast to methods that use dense correspondence ground-truths as direct supervision for local feature matching training, we train 3DG-STFM: a multi-modal matching model (Teacher) to enforce the depth consistency under 3D dense correspondence supervision and transfer the knowledge to 2D unimodal matching model (Student).
Both teacher and student models consist of two transformer-based matching modules that obtain dense correspondences in a coarse-to-fine manner.
The teacher model guides the student model to learn RGB-induced depth information for the matching purpose on both coarse and fine branches.
We also evaluate 3DG-STFM on a model compression task.
To the best of our knowledge, 3DG-STFM is the first student-teacher learning method for the local feature matching task.
The experiments show that our method outperforms state-of-the-art methods on indoor and outdoor camera pose estimations, and homography estimation problems.
Code is available at:\href{https://github.com/Ryan-prime/3DG-STFM}{https://github.com/Ryan-prime/3DG-STFM}.

\end{abstract}

\section{Introduction}
\label{sec:intro}

Establishing correspondences between overlapped images is critical for many computer vision tasks including structure from motion (SfM), simultaneous localization and mapping (SLAM), visual localization, etc. 
Most existing methods that tackle this problem follow the classical tri-stage pipeline, i.e., feature detection~\cite{dog,fast}, feature description~\cite{sift,brisk,surf,deepcd,superpoint,magicpoint}, and feature matching~\cite{sift,NCN,superglue}.
To improve efficiency, HLoc~\cite{hloc} was proposed to incorporate these matching techniques for visual localization.
Several recent works~\cite{NCN,efficient_NCN,dual,loftr} attempted to avoid the detection step and established a dense matching by considering all points from a regular grid.
These dense matching approaches aim to supply interest points in low-texture regions and provide sufficient candidates for the matching purpose.

To generate dense ground-truth correspondences as supervision, depth maps, camera intrinsic and extrinsic matrices are used for the calculation of point reprojections from one image to the other~\cite{superglue,loftr,dual}.
Although photometric objective, widely used in optical flow estimation~\cite{optical1,optical2,optical3}, could provide dense correspondences, its constant brightness assumption is not allowed to be generalized for the geometric matching problem. One typical adversarial scenario is image pairs taken under radically different illumination.
On the other hand, given a set of images with dense correspondences, triangulation could easily reconstruct the 3D scene and depth maps.
Therefore, depth information is implicitly provided by dense correspondence supervision.

\begin{figure}[t]
  \centering
  \begin{subfigure}{0.49\linewidth}
    \centering
    \includegraphics[width=0.75\linewidth]{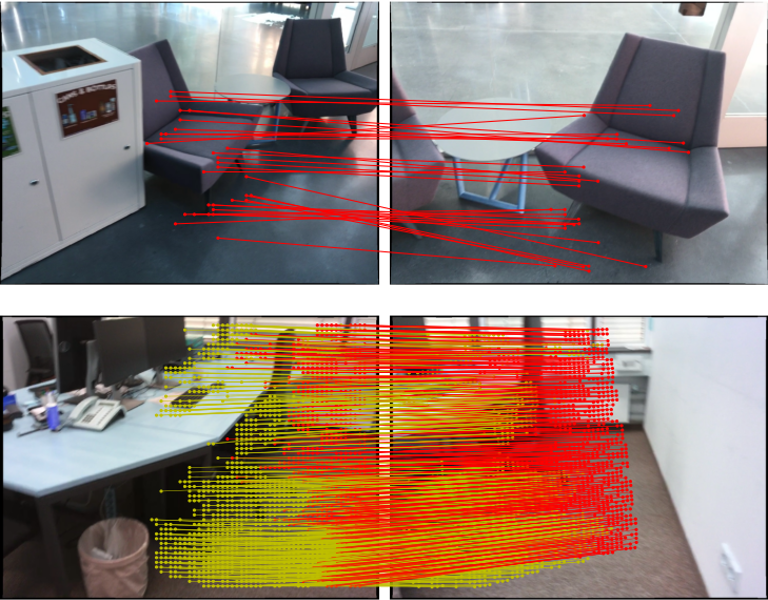}
    \caption{LoFTR}
    \label{fig:loftr}
  \end{subfigure}
  \hfill
  \begin{subfigure}{0.49\linewidth}
    \centering
    \includegraphics[width=0.75\linewidth]{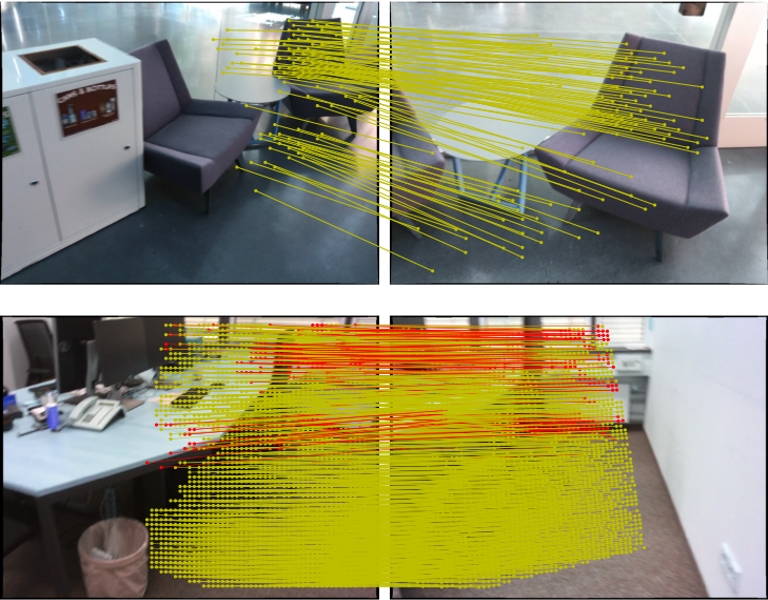}
    \caption{3DG-STFM}
    \label{fig:our}
  \end{subfigure}
  \caption{\textbf{Comparison between dense local feature matching method LoFTR~\cite{loftr} and the proposed method 3DG-STFM.}
  This example demonstrates that our approach, embedded the depth distribution via student-teacher learning, could find the correct correspondences under challenging scenario with repetitive patterns and low-texture regions. The red color indicates epipolar error beyond $5\times10^{-4}$ (in the normalized image coordinates).}
  \label{fig:intro}
\end{figure}

However, to the best of our knowledge, none of the existing methods explored the depth modality distribution during the training phase.
Depth maps, unlike RGB images, provide 3D information, which depicts the geometry distribution in an explicit manner. 
We argue that the introduction of depth modality distribution can provide two-fold benefits.
First, depth information, even if in lower quality or sparse, can remove lots of ambiguity in 2D image space and enforce geometric consistency for feature matching, which is very difficult using only RGB inputs. 
That is particularly true when there are multiple similar objects within the image pair. In that case, most of existing methods tend to find implausible matching candidates since they purely discriminate 2D descriptors without depth or size knowledge.
An example is shown in the first row of Fig.~\ref{fig:intro}, where the baseline method is confused by the similar 2D appearance and incorrectly matches the closer chair to the further one. 
Second, as the example shown in the second row of Fig.~\ref{fig:intro}, low texture area of single object haunts 2D descriptor in terms of enforcing dense and consistent matching. That deficiency can also be nicely regularized by leveraging the discrimination of depth modality.

Despite the advantage of depth information, high quality RGB-D inputs can only be collected in well-controlled lab environment, and very few, especially low cost devices can capture similar well aligned RGB-D pairs in real world scenarios. 
Most imaging systems are only equipped with RGB sensors as input and cannot afford high computational cost stemmed from multi-modal inference. 
That makes naive multi-modal fusion of RGB and depth inputs during both inference and training a restrictive solution. Consequently, a good way of transferring expensive RGB-D knowledge into RGB modality inference is needed in practical scenarios, considering constraints from both hardware and computational load.

Motivated by these observations, we propose 3DG-STFM, a student-teacher learning framework, to transfer depth knowledge learned by a multi-modal teacher model to a unimodal student model to improve the local feature matching.
To the best of our knowledge, 3DG-STFM is the first student-teacher learning architecture to transfer cross-modal knowledge on the image matching problem. 
The method aims to find the depth and RGB correlational distribution in RGB-D images and transfer the knowledge to the RGB student branch by maintaining such distribution. 
Therefore, depth modality is not explicitly required in the actual inference process (student branch). 

We propose attention mechanisms to guide the student model to study the teacher model's matching distribution and learning priority. 
Therefore, with RGB images as input, the student unimodal model could explore  RGB-induced depth information and learn multi-modal matching strategies.
The main contributions of this paper are summarized as follows:
\begin{itemize}
  \item We propose the first student-teacher learning architecture on the local feature matching problem that learns the induced depth distribution distilled from dense RGB-D correspondence supervision.
  \item We propose attentive knowledge transfer strategies to help the student model understand the matching distribution and learning priority during the training instead of learning point-to-point matching.
  \item  We show that the proposed model produces high-quality dense correspondences on a range of matching tasks and achieves state-of-the-art results on both camera pose and homography estimation tasks.
\end{itemize}

\section{Related Work}
\subsection{Learning-based Dense Local Feature Matching}
In the past decades, many groups made great efforts to improve the local feature matching pipeline, i.e., feature detection, feature description and feature matching, and achieved promising performance by leveraging learning-based techniques.
DeTone et al. proposed Superpoint~\cite{superpoint}, a self-supervised learnable interest point detector and descriptor. 
ViewSynth~\cite{viewsynth} designed a depth map keypoint detection method without using RGB domain information.
Instead of learning better task-agnostic local features, SuperGlue~\cite{superglue} built a densely connected graph between two sets of keypoints by leveraging a Graph Neural Network (GNN). 
Geometric correlation of the keypoints and their visual features are integrated and exchanged within the GNN using the self and cross attention mechanism.
However, those detector-based local feature matching algorithms only produced sparse keypoints, especially in low-texture regions.

To address the above problem, detector-free methods~\cite{efficient_NCN,dual,loftr,cotr} proposed pixel-wise dense matching methods.
~\cite{ucn} and ~\cite{learning_dense_2016} used contrastive loss to learn dense feature descriptors and were followed by the nearest neighbor search for the matching purpose.
NCNet~\cite{NCN} proposed an end-to-end approach by directly learning the dense correspondences.
It enumerated all possible matches between two images and constructed a 4D correlation tensor map. The 4D neighborhood consensus networks learned to identify reliable matching pairs and filtered out unreliable matches accordingly.
Based on this concept, Sparse NCNet~\cite{efficient_NCN} improved NCNet's efficiency and performance by processing the 4D correlation map with submanifold sparse convolutions~\cite{sparseconv}.
And DRC-Net~\cite{dual} proposed a coarse-to-fine approach to generate higher accuracy dense correspondences.
Recently, LoFTR~\cite{loftr} was proposed to learn global consensus between image correspondences by leveraging Transformers.
Inspired by~\cite{superglue}, the attention mechanism was used to learn the mutual relationship among features.
For memory efficiency, the coarse matching features were first predicted and then fed to a small transformer to produce the final fine-level matches.
Benefiting from the global receptive field of Transformers, LoFTR improved the matching performance by a large margin.
All above mentioned dense local feature matching approaches needed dense ground-truth correspondences as supervision.
None of the dense matching methods has explored any modality beyond 2D image space in which the feature ambiguities often exist due to the missing information in depth. 

\subsection{Student-Teacher Learning}
Student-teacher learning has been actively studied in knowledge transfer context including model compression~\cite{related2,KD}, acceleration~\cite{related8,related9}, and cross-modal knowledge transfer~\cite{cross_modal_KD,cross_modal_KD2}.
Given a well-trained teacher model with large weight, the goal of the student-teacher learning is to distill and compress the knowledge from the teacher, and guides the lightweight student model for better performance.
On the other hand, data with multiple modalities commonly provides more valuable supervisions than single modality data and could benefit model performance.
However, due to the lack of data or labels for some modalities during training or testing, it is important to transfer knowledge between different modalities.

Due to different network architectures, many different knowledge transfer approaches have been proposed.
The most popular response-based knowledge for image classification was Knowledge Distillation (KD) loss proposed by~\cite{KD}.
In this method, KD loss employed the distribution of neural response of the last output layer, logits layer, of the teacher model and guided the student to learn the distribution.
Besides the output, the intermediate layer's feature representation was also used to train the student model~\cite{fitnets}.
Zagoruyko et al.~\cite{attention_distill} proposed a method to transfer the attention instead of the feature representations to achieve a better distillation performance. 
And NST~\cite{nst} provided a method to learn a similar activation of the neurons. 
Moreover, there are many other related approaches~\cite{GCN_KD,related1,related3,related4,related6,realted7}. 
However, none of them provided a knowledge transfer solution for correspondence matching problems, which need to consider all mutual relationships among local features of different images. 

\begin{figure}[t]
  \centering
   \includegraphics[width=0.85\linewidth]{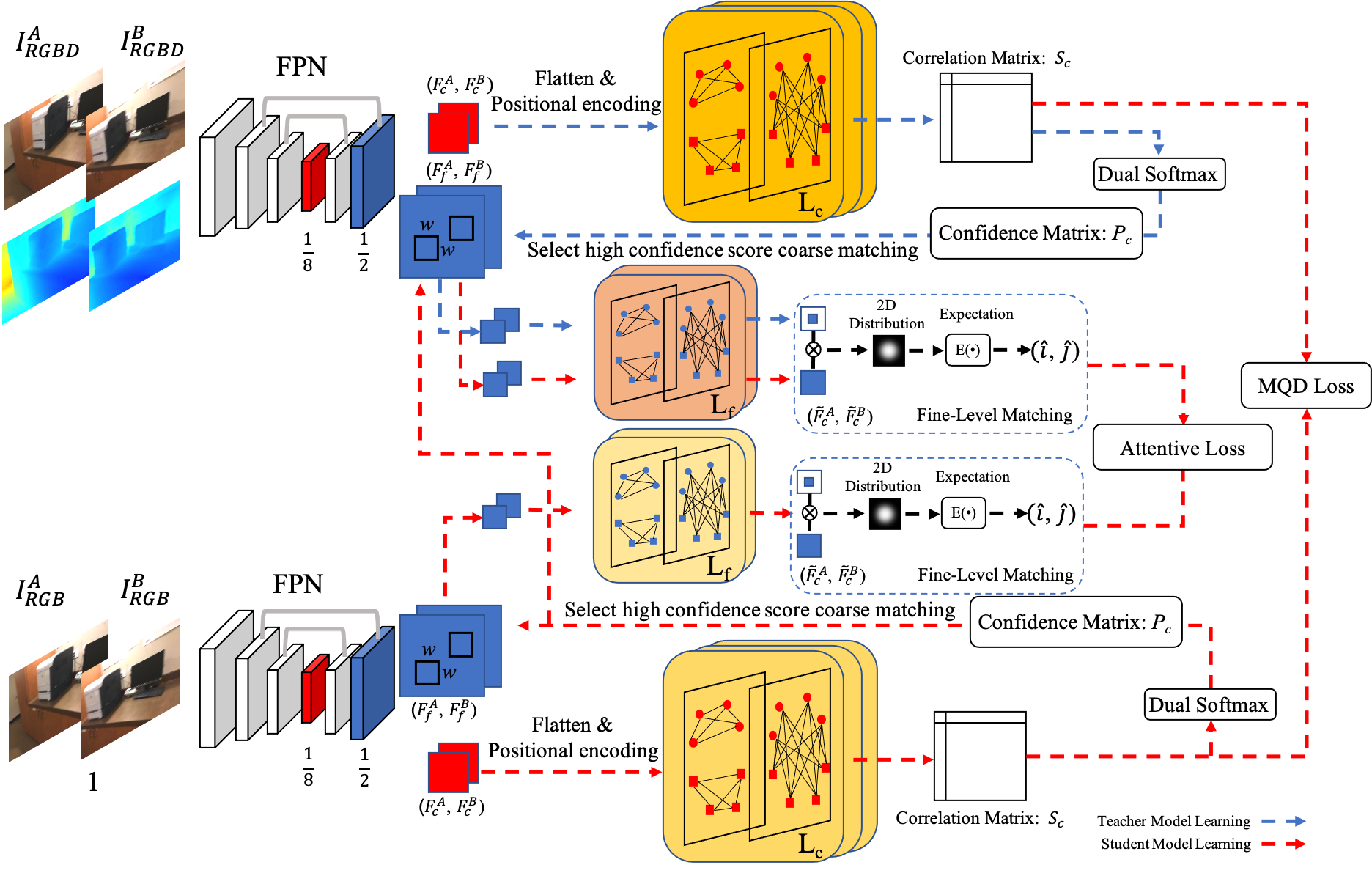}
   \caption{\textbf{Overview of 3DG-STFM.} For each of student or teacher branch, Feature Pyramid Networks (FPN)~\cite{FPN} are used to extract coarse-level local features $(F_{c}^{A},F_{c}^{B})$ and fine-level features $(F_{f}^{A},F_{f}^{B})$ with $\frac{1}{8}$ and $\frac{1}{2}$ of the original image resolution.
   The coarse level transformer consisting of $\mathrm{L_{c}}$ attention layers finds coarse pairs and their matching scores.
   Matches with high confidence scores will be selected and mapped to a fine-level feature map. Surrounding features on $F_{f}^{A}, F_{f}^{B}$ are collected by the $w \times w$ size window and fed to a fine-level transformer with $\mathrm{L_{f}}$ attention layers.
   The fine-level matching module is applied to predict correspondences $(\hat{i}, \hat{j})$ on subpixel-level.
   The teacher model is first trained under direct supervision. During student training, it will be frozen and provide additional supervision via attentive loss and Mutual Query Divergence (MQD) loss.
   }
   \label{fig:overview}
\end{figure}

\section{Method}
Our proposed system, 3DG-STFM, is to train a unimodal local feature matching model (Student) by leveraging the knowledge from a well-trained multi-modal model (Teacher).
As shown in Fig.~\ref{fig:overview}, the RGBD image pairs $(I_{RGBD}^{A}, I_{RGBD}^{B})$ and RGB image pairs $(I_{RGB}^{A}, I_{RGB}^{B})$ are fed to teacher and student branches separately.
The labels of dense correspondences provide direct supervision during the teacher or student training.
Once we reach a well-trained multi-modal teacher model, two strategies are proposed for cross-modal knowledge transfer:
$(1)$ Using the Mutual Query Divergence (MQD) loss guides the student model to learn the coarse-level matching distributions embedded in the teacher model's correlation matrix $S_{c}$.
$(2)$ Using the attentive loss guides the student at the fine-level module to pay more attention to the teacher's confident predictions and learn the matching distribution with priority. 

Our method is based on the matching strategies mentioned in LoFTR~\cite{loftr} due to their high performances. In this section, we will first introduce the transformer-based model in Section~\ref{sec:backbone}.
Section~\ref{sec:coarse} and ~\ref{sec:fine} will describe our knowledge transfer strategies over both coarse and fine levels.

\subsection{Transformer-based Local Feature Matching}\label{sec:backbone}
As shown in Fig.~\ref{fig:overview}, two transformer-based matching modules, inspired by~\cite{loftr}, are adopted in both teacher and student branches of our 3DG-STFM system.

\begin{figure}[t]
  \centering
    \includegraphics[width=0.5\linewidth]{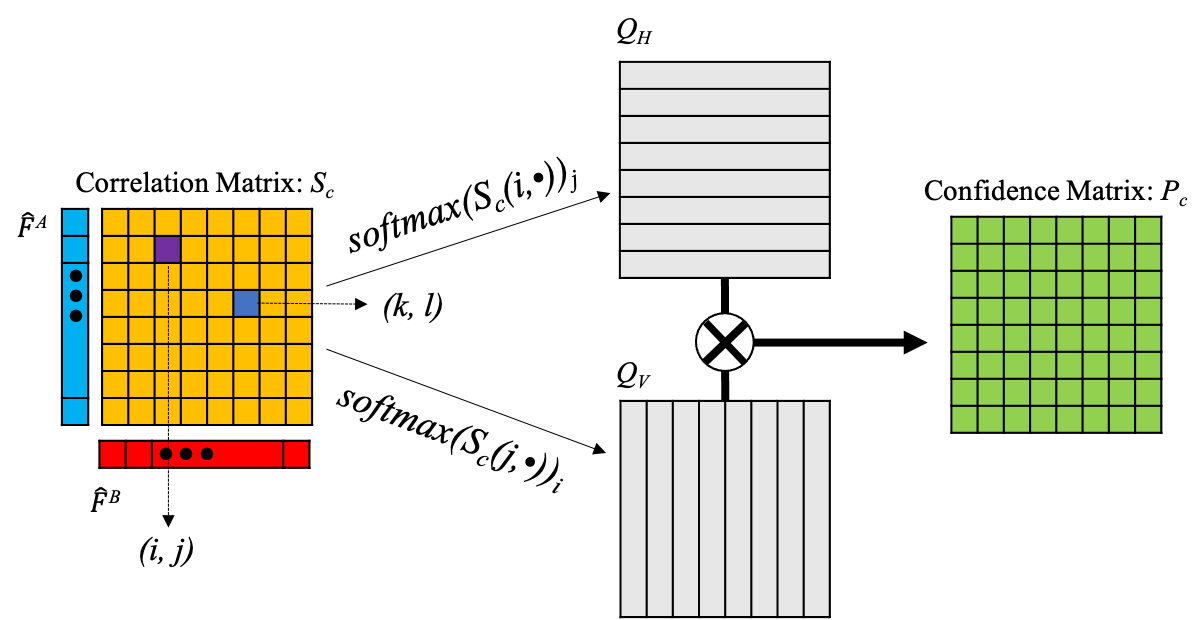}
    \caption{\textbf{Coarse-level differentiable matching mechanism.} Transformer outputs are correlated to generate correlation matrix $S_{c}$. Dual-softmax~\cite{disk,NCN} operation is applied on two dimensions to obtain the matching probability.}
    \label{fig:coarse1}
\end{figure}

\noindent\textbf{Coarse-level Matching.} 
Given the coarse-level feature map in dimension $h\times w \times c$, we flatten them into $hw \times c$ and do the positional encoding~\cite{DETR}.
The encoded local feature vector will be fed to a coarse-level matching transformer.
Unlike classical vision transformer~\cite{TR_cls,TR_det,TR_seg} focusing on self-attention, the matching transformer adds a cross-attention layer to consider the relations between pixels from different images. 
We interleave the self-attention and cross-attention layers in matching transformer modules by $\mathrm{L_{c}}$ times.
As shown in Fig.~\ref{fig:coarse1}, the output of the coarse matching transformer $\{\hat{F}^{A},\hat{F}^{B}\}$ corresponding to two different images $\{I^{A}, I^{B}\}$ will be used to calculate correlation matrix $S_{c}$ by $S_{c}(i,j) = Corr(\hat{F}_{i}^{A},\hat{F}_{j}^{B})$, in which $\hat{F}_{i}^{A}$ and $\hat{F}_{j}^{B}$ indicate local feature at position $i$ of $I^{A}$ and local feature at position $j$ of $I^{B}$.
The dual-softmax~\cite{disk,NCN}, two softmax operations with temperature $\tau = 0.1$ in horizontal and vertical directions, is applied on the correlation matrix to calculate forward and backword matching probability:
$P_{A\rightarrow B}(i,j) = softmax(\frac{1}{\tau}S_{c}(i,\cdot))_{j}$ and 
$P_{A\leftarrow B}(i,j) = softmax(\frac{1}{\tau}$\\$S_{c} (\cdot, j))_{i}$.
The confidence matrix $P_{c}$ with the final matching probabilities has same dimension as $S_{c}$ and is calculated by:
$P_{c}(i,j) = P_{A\rightarrow B}(i,j) \cdot P_{A\leftarrow B}(i,j)$.
We call the output of horizontal and vertical softmax as query matrix $\{Q_{H}, Q_{V}\}$ since they depict query results of each feature from one image to another and vice versa.
Given the ground-truth matrix derived from correspondence labels, we calculate the cross-entropy loss:
\begin{equation}
  \mathcal{L}_{c} = -\frac{1}{|\mathcal{M}_{gt}|}\sum_{(i,j)\in \mathcal{M}^{gt}_{c}} FL(P_{c}(i,j)) \log P_{c}(i,j)
  \label{eq:LC}
\end{equation}
\begin{equation}
  FL(p) = \alpha(1-\hat{p})^{\gamma}, \hat{p}=  \left\{ \begin{array}{rl}
    p & \mbox{if y=1}\\
    1-p & \mbox{otherwise} 
  \end{array}\right.
  \label{eq:FL}
\end{equation}
in which $P_{c}$ is the confidence matrix and $\mathcal{M}_{gt}$ is the correspondence set generated by ground-truth labels.
We follow~\cite{loftr} to set a focal loss term, $FL$ with predicted probability $p$, to address the imbalance between matching and non-matching pairs.

\noindent\textbf{Fine-level Matching.}
Based on the confidence matrix $P_{c}$, matching pairs with probability scores higher than a threshold $\theta_{c}$ are selected and refined by a fine-level matching module.
The selected coarse-level features are upsampled and concatenated to fine-level features cropped by $w \times w$ size windows before passing to the fine-level matching transformer.

The fine-level matching transformer is a lightweight transformer containing $\mathrm{L}_{f}$ attention layers.
It aggregates the contextual information to generate features $\{\Tilde{F}_{f}^{A}, \Tilde{F}_{f}^{B}\}$ and passes them to a differentiable matching module.
Instead of generating a confidence matrix, the fine-level matching module selects the center feature of $\Tilde{F}_{f}^{A}$ and correlates with all features in $\Tilde{F}_{f}^{B}$.
The similarity distribution is generated and the expectation $\mu$ is treated as the prediction.
The final loss based on direct supervision is calculated by:
\begin{equation}
  \mathcal{L}_{f} = \frac{1}{|\mathcal{M}_{f}|} \sum_{(\hat{i},\hat{j})\in \mathcal{M}_{f}}\frac{1}{\sigma^{2}(\hat{i})}||\mu(\hat{i})-\hat{j}_{gt}||_{2}^{2}
  \label{eq:finelevel}
\end{equation}
where $\hat{j}_{gt}$ is the ground-truth position we wrap from image solution to fine-level heatmap scale. $\mu(\hat{i})$ is the prediction associated to coarse position $\hat{i}$ and $\sigma^{2}(\hat{i})$ is the total variance of heatmap distribution. $\mathcal{M}_{f}$ is the set of fine matches predicted by module.
The total variance of the similarity distribution is treated as uncertainty to assign a weight to each fine-level match.
The larger total variance indicates it is an uncertain prediction and associate with low weights.

\begin{figure}[t]
  \centering
  \begin{subfigure}{0.49\linewidth}
    \centering
    \includegraphics[width=1\linewidth]{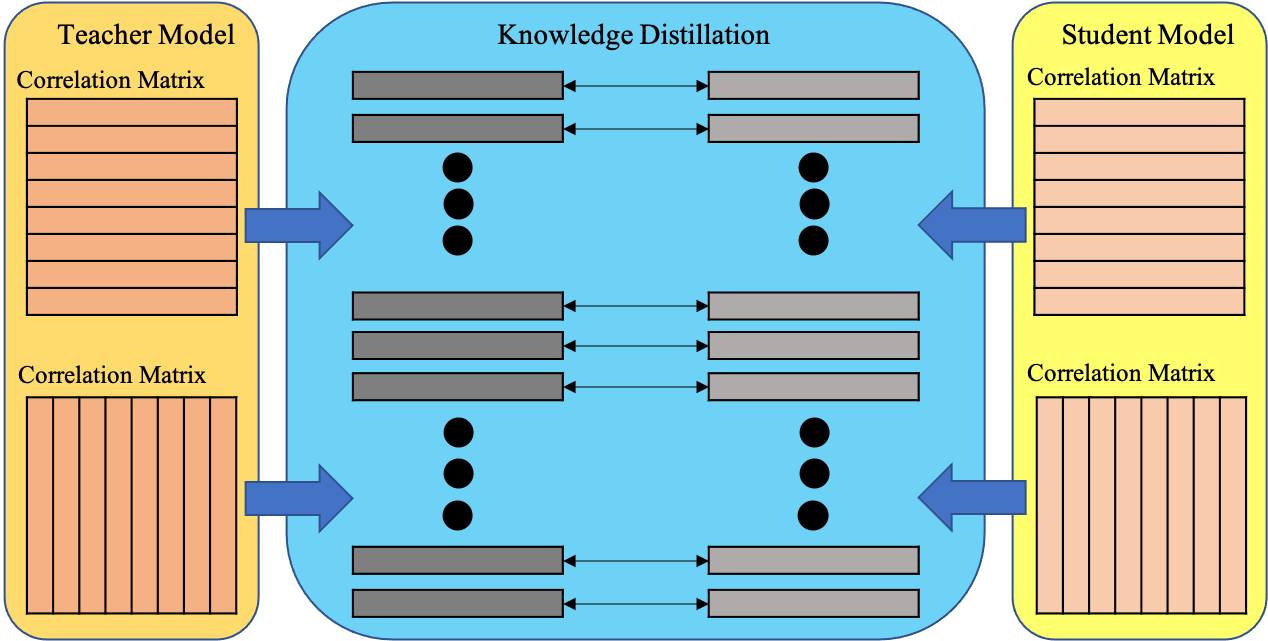}
    \caption{Coarse-level knowledge transfer.}
    \label{fig:coarse2}
  \end{subfigure}
  \hfill
  \begin{subfigure}{0.49\linewidth}
   \centering 
    \includegraphics[width=0.78\linewidth]{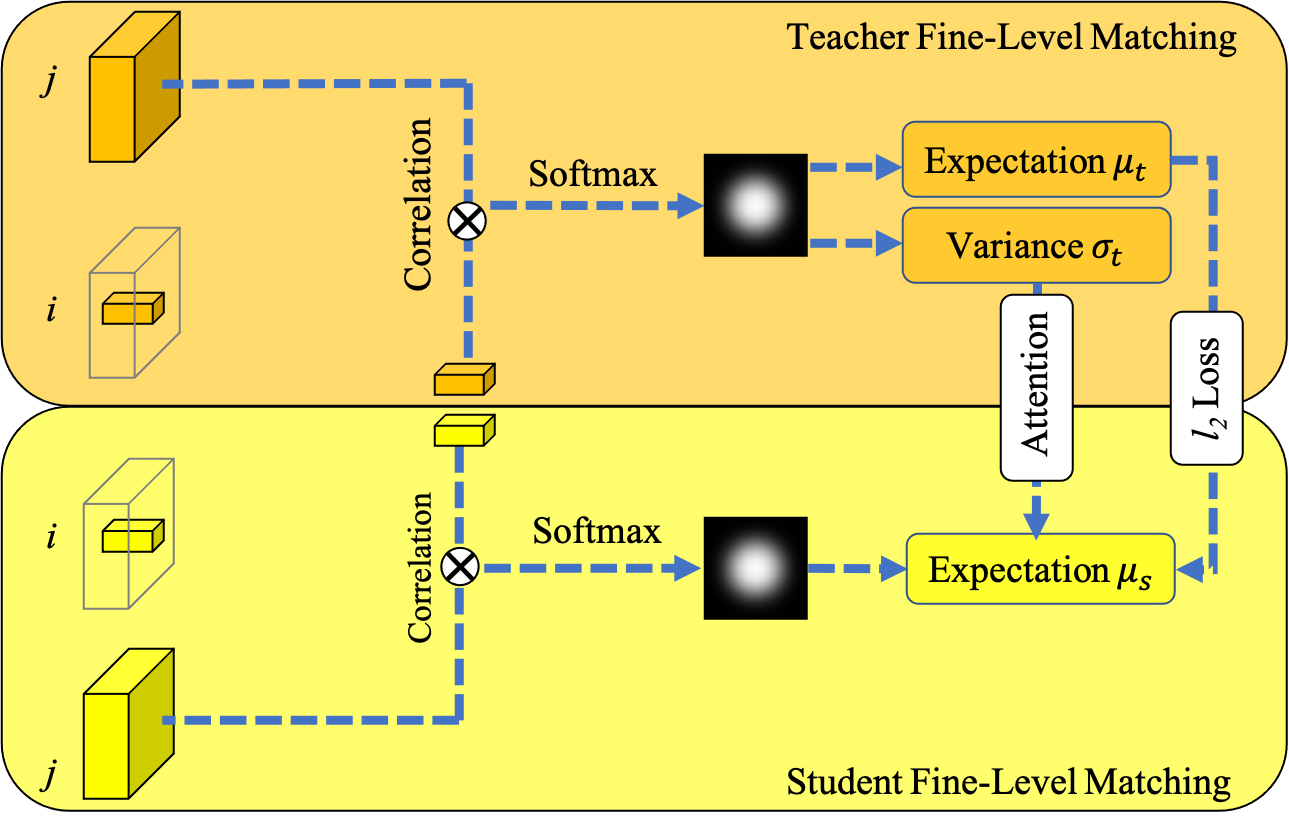}
    \caption{Fine-level attentive knowledge transfer}
    \label{fig:fine}
  \end{subfigure}
  \caption{\textbf{Knowledge transfer on coarse-level and fine-level.} (a) The correlation matrix is decomposed into multiple independent distributions depicting mutual query processes for the student learning.(b) One center point of a fine-level feature is selected and correlates with all points of the other feature map for heatmap distribution generation. Both expectation and variance of teacher branch's heatmap are used for fine-level knowledge transfer.
  }
  \label{fig:coarse_fine}
\end{figure}

\subsection{Coarse-level Knowledge Distillation}\label{sec:coarse}
A response-based knowledge distillation strategy is applied to help the student learn from the teacher on a coarse level.
This method distills the logits layer's distribution and guides the student to learn.
As aforementioned, the logits layer's output in our case is correlation matrix $S_{c}$ with size $hw \times hw$.
Each row or column depicts the relation between one pixel and each pixel of the other image.
The dual softmax operation could be treated as a query process in two directions.
Local features at position $i\in F_{c}^{A}$ retrieve the closest feature from all positions $j \in F_{c}^{B}$ and vice versa.
Many existing response-based knowledge \cite{KD,cross_modal_KD} distillation methods treat the logits layer's output as a single distribution.
However, as shown in Fig.~\ref{fig:coarse1}, exploring the relationship between $S_{c}(i,j)$ and $S_{c}(k,l)$ in different row and column is meaningless for matching purpose.
These uninterpretable relations could produce extra loss that confuses the knowledge transfer process.

Instead of learning a single distribution from the correlation matrix $S_{c}$ in teacher, we split the distribution into two matching query matrices, as shown in Fig.~\ref{fig:coarse2}, to avoid the unpredictable correlations between them. 
Based on Knowledge Distillation (KD) loss~\cite{KD}, we propose Mutual Query Divergence loss $\mathcal{L}_{MQD}$ that employs all $2 \times hw$ mutual query distributions:
\begin{equation}
  \mathcal{L}_{MQD} = \frac{1}{n}\big(-\sum_{i=1}^{n}FL(p_{S}^{(i)})\hat{p}^{(i)}_{S}\log(\hat{p}^{(i)}_{T})\big)
  \label{eq:kdloss}
\end{equation}
\begin{equation}
  p^{(l)}_{S} =  \frac{\exp(o_{S}^{(k)})}{\sum_{k=1}^{L}\exp(o_{S}^{k})},
\hat{p}^{(l)}_{S} =  \frac{\exp(\frac{o_{S}^{(k)}}{T})}{\sum_{k=1}^{L}\exp(\frac{o_{S}^{k}}{T})},
  \hat{p}^{(l)}_{T} = \frac{\exp(\frac{o_{T}^{(k)}}{T})}{\sum_{k=1}^{L}\exp(\frac{o_{T}^{k}}{T})}
  \label{eq:kdloss1}
\end{equation}
in which $\hat{p}_{S}^{l}$ and $\hat{p}_{T}^{l}$ are student and teacher's query distributions distilled from their logits layer's outputs $o_{S}$ and $o_{T}$ at temperature $T$.
Additional focal loss weight $FL$ (Equation~\ref{eq:FL}) is added to balance the matching/unmatching ground-truth pairs.
$p_{S}^{l}$ is the standard confidence score predicted by student model expressed in  Equation~\ref{eq:kdloss1}.
The $\mathcal{L}_{MQD}$ on coarse level is the mean of KD loss of all $n$ distribution, where $n$ is equal to $2 \times hw$ in our case.
Based on this loss, the coarse level matching module pays attention to the distributions benefit matching and ignores noisy information across distributions.

\subsection{Fine-level Attentive Knowledge Transfer}\label{sec:fine}
After transferring the coarse level matching knowledge from the teacher model with the mutual query distribution distillation, an attentive loss ($\mathcal{L}_{att}$) is proposed for the student's fine-level matching module.
Instead of learning point-to-point matching under the supervision of ground-truth, $\mathcal{L}_{att}$ explores the matching distribution and learning priority of the teacher model.

As shown in Fig.~\ref{fig:fine}, the fine-level local feature matching is based on the differentiable matching approach that could produce a heatmap that represents the matching probability of each pixel in the neighborhood of $j$ with $i$. 
By computing expectation $\mu$ over the probability distribution, we get the final position $\hat{j}$ with sub-pixel accuracy on $I^{B}$.
The uncertainty of the prediction is also measured by the total variance of the correlation distribution.
During the student-teacher learning process, both branches could generate heatmaps.
We treat heatmaps of teacher model and student model as gaussian distributions $\mathcal{N}_{t}(\mu_{t},\sigma^{2}_{t})$ and $\mathcal{N}_{s}(\mu_{s},\sigma^{2}_{s})$.
The  Kullback–Leibler (KL) divergence loss ($\mathcal{L}_{KL}$) is applied to help the student learn the distribution from the teacher.
The KL divergence of two gaussian distributions could be written as:
\begin{align}
    \mathcal{L}_{KL}(\mathcal{N}_{s},\mathcal{N}_{t}) = \log(\frac{\sigma_{t}}{\sigma_{s}})+\frac{\sigma^{2}_{s}+(\mu_{s}-\mu_{t})^{2}}{2\sigma_{t}^{2}}-\frac{1}{2}
  \label{eq:KL} 
\end{align}

Although total variance $\{\sigma_{t}, \sigma_{s}\}$, and $\sigma(\hat{i})$ (Equation~\ref{eq:finelevel}) are included in the loss, the optimizer would decrease loss by increase the total variance.
To avoid the incorrect loss, the gradient is not backpropagated through $\sigma_{s}$,  $\sigma_{t}$, and $\sigma(\hat{i})$. 
Therefore, we generate $\mathcal{L}_{att}$ by removing those constant variable of $\mathcal{L}_{KL}$:
\begin{equation}
  \mathcal{L}_{att} = \frac{1}{|\mathcal{M}_{f}|}\sum_{(\hat{i},\hat{j})\in \mathcal{M}_{f}}\frac{(\mu_{s}^{(\hat{i})}-\mu_{t}^{(\hat{i})})^{2}}{2\sigma_{t}^{(\hat{i})2}}
  \label{eq:attloss}
\end{equation}
where the $\mu_{s}^{(\hat{i})}$ and the $\mu_{t}^{(\hat{i})}$ are the expectations of student's and teacher's output distributions which corresponding to match $(\hat{i}, \hat{j})$ in fine-level correspondence set $\mathcal{M}_{f}$. 
Therefore, the total loss is a mean of the weighted sum of all the fine-level pairs' $l_{2}$ loss in matching set $\mathcal{M}_{f}$.
We call this attentive loss since it could be treated as a $l_{2}$ distance loss that pays more attention to the prediction associated with large attention weight $\frac{1}{2\sigma_{t}^{2}}$.
The total variance is commonly treated as a metric for certainty measure.
The teacher prediction with a small total variance indicates the teacher is quite certain about the location of the correspondence.
In this case, the loss is assigned with a large weight to guide the student model to learn those certain predictions from the teacher in priority.

\subsection{Supervision}
Both teacher and student training processes are under the direct supervision provided by correspondence ground-truths.
The teacher model provides extra supervision during the student model training.
For direct supervision, we follow the same procedure mentioned in ~\cite{superglue,NCN,loftr} that uses the camera intrinsic, extrinsic matrices, and depth maps to compute the dense correspondences.
To supervise coarse-level matching training, mutual nearest neighbors of the two sets of $\frac{1}{8}$-resolution grids are selected as ground-truth $\mathcal{M}^{gt}_{c}$.
The pixel-level matching positions could be used for $l_{2}$ loss and supervise the fine-level matching learning.
The final loss for the teacher and student model is:
\begin{equation}
  \mathcal{L}_{teacher} = \lambda_{0}\mathcal{L}_{c}+\lambda_{1}\mathcal{L}_{f}
  \label{eq:teacher}
\end{equation}
\begin{equation}
  \mathcal{L}_{student} = \lambda_{0}\mathcal{L}_{c}+\lambda_{1}\mathcal{L}_{f}+\lambda_{2}\mathcal{L}_{MQD}+\lambda_{3}\mathcal{L}_{att}
  \label{eq:student}
\end{equation}
in which $\mathcal{L}_{c}$ and $\mathcal{L}_{f}$ are coarse-level and fine-level loss under direct supervision described in Equation~\ref{eq:LC} and Equation~\ref{eq:finelevel}.
The student model is also guided by the teacher model via Mutual Query Divergence loss $\mathcal{L}_{MQD}$ and attentive loss $\mathcal{L}_{att}$ for the coarse and the fine level knowledge transfer.

\subsection{Implementation Details}\label{sec:implement}
We train the indoor model of 3DG-STFM on the ScanNet~\cite{scannet} dataset and the outdoor model on the MegaDepth~\cite{megadepth} dataset.
The coarse-level transformer contains 4 attention layers, and the fine-level transformer has 1 attention layer.
Each attention layer consists of a self-attention and a cross-attention layer with 8 heads.
The focal loss parameters $\{\alpha, \gamma\}$ are set as $\{0.25, 2.0\}$.
The confidence score threshold $\theta_{c}$ is set to $0.2$ to remove unreliable correspondences.
The window size $w$ is 5.
For indoor dataset ScanNet, the models are trained using AdamW
with an initial learning rate of $6 \times 10^{-3}$ on 32 2080Ti GPUs.
All images are resized to $640\times 480$.
The weights of losses $\{\lambda_{0}, \lambda_{1}, \lambda_{2}, \lambda_{3}\}$ are set as $\{0.25, 0.25, 4.0, 0.25\}$. 
The outdoor models for Megadepth are trained using AdamW
with an initial learning rate of $8 \times 10^{-3}$ on 16 P100 GPUs.
The weights of losses $\{\lambda_{0}, \lambda_{1}, \lambda_{2}, \lambda_{3}\}$ are set as $\{0.25, 0.25, 1.0, 0.25\}$.
It is worth mentioning that our method is based on LoFTR~\cite{loftr}, which provides two version implementations for the outdoor dataset in their official code. 
One is for $840\times 840$ resolution image and consumes 24 GB RAM during the training. The other is training on $640\times 640$ image pairs and feasible for 16 GB RAM GPUs.
In this work, we treat the latter one as the baseline for the outdoor pose estimation and homography estimation tasks, and our 3DG-STFM is also trained on images resized to $640\times 640$ with padding.
We normalize depth maps in both ScanNet and Megadepth in the training process. 
The depth maps of ScanNet are in the range of 0 to 10 meters. We normalize it to $[0, 1]$ and concatenate it to RGB images for multi-modal training. 
On the other hand, Megadepth's depth maps are relative estimations that come from COLMAP~\cite{colmap} reconstructions and have a pretty large range.
We normalized them to $[0,1]$ for each pair of images for teacher model training.
\section{Experiments}
\subsection{Indoor Pose Estimation}
\noindent \textbf{Dataset.} We use ScanNet~\cite{scannet}, a large-scale indoor scene dataset composed of 1613 monocular sequences with depth maps and camera poses. 
This dataset is quite challenging due to extensive texture-less regions and repetitive patterns.
Following the~\cite{superglue,loftr}, we sample 230M image pairs with overlap scores between 0.4 and 0.8 for training and the student model is evaluated on the 1500 testing pairs.
The images are resized to $640 \times 480$ to fit the depth map's dimension.

\noindent\textbf{Evaluation Protocol.}
Following~\cite{loftr}, we report the AUC
of the pose error at thresholds $(5^{\circ}, 10^{\circ}, 20^{\circ})$.
The pose error is defined as the maximum of angular error in rotation and translation.
The predicted matches are used to solve the essential matrix with RANSAC.

\setlength{\tabcolsep}{4pt}
\begin{table}[t]
  \centering
  \caption{\textbf{Evaluation on ScanNet~\cite{scannet} for indoor pose estimation.} The AUC of the pose error in percentage is reported.}
  \begin{tabular}{@{}ccccc@{}}
    \hline
    \multirow{2}{*}{Category} & \multirow{2}{*}{Method} & \multicolumn{3}{c}{Pose estimation AUC}\\\cline{3-5}
    \multirow{2}{*}{} & \multirow{2}{*}{} &$@5^{\circ}$ &$@10^{\circ}$ &$@20^{\circ}$\\
    \hline
    Multi-Modal & 3DG-STFM Teacher &27.93&47.11&63.74\\
    \hline
    \multirow{8}{*}{Detector-based} 
    &ORB~\cite{orb}+GMS~\cite{gms} &5.21&13.65&25.36\\
    &D2-Net~\cite{d2net}+NN &5.25&14.53&27.96\\
    &ContextDesc~\cite{contextdesc}+Ratio Test~\cite{sift} &6.64&15.01&25.75\\
    &SP~\cite{superpoint}+NN &9.43&21.53&36.40\\
    &SP~\cite{superpoint}+PointCN~\cite{pointCN} &11.40&25.47&41.41\\
    &SP~\cite{superpoint}+OANet~\cite{oanet} &11.76&26.90&43.85\\
    &SP~\cite{superpoint}+SGMNet~\cite{sgmnet} &15.40&32.06&48.32\\
    &SP~\cite{superpoint}+SuperGlue~\cite{superglue} &16.16&33.81&51.84\\
    \hline
    \multirow{3}{*}{Detector free}
    &LoFTR~\cite{loftr} &22.06&40.80&57.62\\
    & 3DG-STFM Student &\textbf{23.58}&\textbf{43.60}&\textbf{61.17}\\
  \hline
  \end{tabular}
  \label{tab:scannet}
\end{table}
\setlength{\tabcolsep}{1.4pt}

\noindent\textbf{Results.}
Since the released DRC-Net is trained on MegaDepth and LoFTR is proved to have better performance, we only consider LoFTR as the state-of-the-art for comparison.
The results in Table~\ref{tab:scannet} show that our student model learns from the teacher model and outperforms all unimodal competitors.
For detector free methods, our student model outperforms LoFTR by $\sim3\% $ at AUC$@10^{\circ}$.
\begin{figure*}[t]
  \centering
    \includegraphics[width=0.9\linewidth]{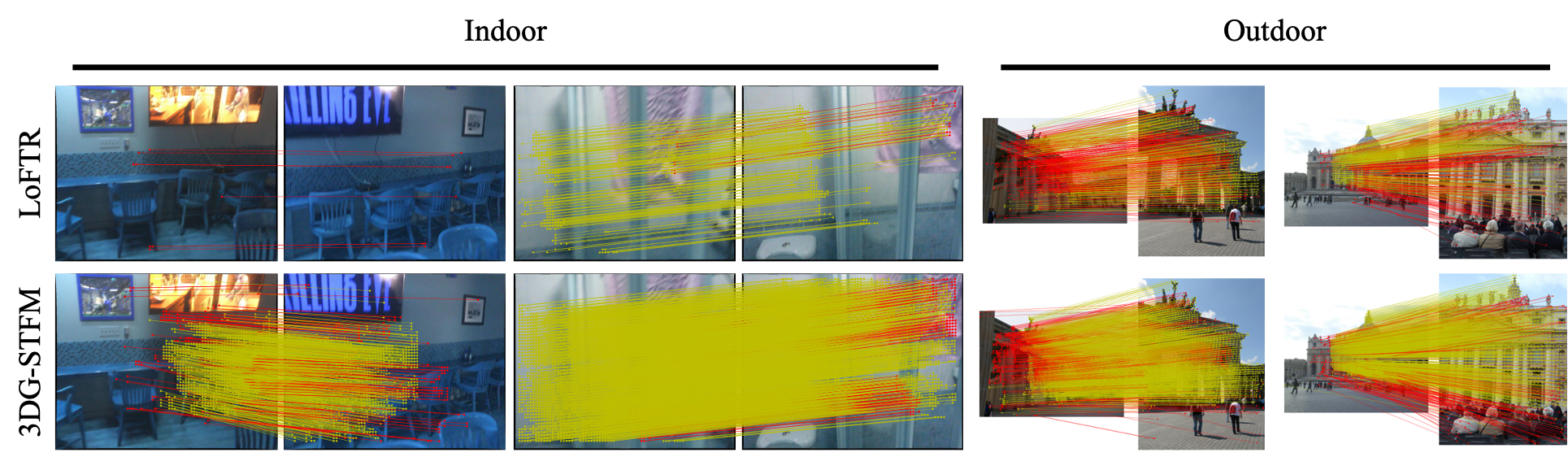}
  \caption{\textbf{Qualitative results.} Our student model is compared to LoFTR~\cite{loftr} in indoor and outdoor scenes. Our method performs better in challenge scenarios with repetitive pattern and low texture region. The red color indicates epipolar error beyond $5\times10^{-4}$ for indoor scenes and $1\times 10^{-4}$ for outdoor scenes (in the normalized image coordinates). More qualitative results in the supplementary.} 
  \label{fig:comparision}
\end{figure*}

For the visual comparison in  Fig.~\ref{fig:comparision}, our student model shows denser and more reliable correspondences than LoFTR does, especially in regions with repetitive patterns.
In addition, our model provides more robust correspondences in the low-texture region, which also benefits the pose estimation.
On average, our student model detects \textbf{1192.84} inlier (epipolar error less than $5\times10^{-4}$ in the normalized image coordinates) correspondences on each pair of indoor images, which is much higher than \textbf{887.04} inlier correspondences of LoFTR.
Both numeric and qualitative results demonstrate the effectiveness of our student model that learns the RGB-induced depth distribution from the teacher model.

\subsection{Outdoor Pose Estimation}\label{sec:outdoor}
\noindent\textbf{Dataset.}
We use MegaDepth~\cite{megadepth}, a dataset consisting of 1M internet images of 196 different outdoor scenes, for outdoor pose estimation evaluation.
We follow DISK~\cite{disk} to select 
1500 pairs for validation.

\begin{table}[t]
  \centering
   \setlength{\tabcolsep}{4pt}
  \caption{\textbf{Evaluation on MegaDepth~\cite{megadepth} for outdoor pose estimation.} The AUC of the pose error in percentage is reported.}
  \begin{tabular}{ccccc}
    \hline\noalign{\smallskip}
    \multirow{2}{*}{Category} & \multirow{2}{*}{Method} & \multicolumn{3}{c}{Pose estimation AUC}\\\cline{3-5}
    \multirow{2}{*}{} & \multirow{2}{*}{} &$@5^{\circ}$&$@10^{\circ}$ &$@20^{\circ}$\\
    \hline\noalign{\smallskip}
    \multirow{1}{*}{Multi-Modal} 
    &3DG-STFM Teacher &53.43&69.81&81.79\\
    \hline\noalign{\smallskip}
    \multirow{1}{*}{Detector-based} 
    &SP~\cite{superpoint}+SuperGlue~\cite{superglue} &42.18&61.16&75.96\\
    \hline\noalign{\smallskip}
    \multirow{3}{*}{Detector free}
    & DRC-Net~\cite{dual} &27.01&42.96&58.31\\
    & LoFTR~\cite{loftr} &51.38&67.11&79.29\\
    & 3DG-STFM Student &\textbf{52.58}&\textbf{68.46}&\textbf{80.04}\\
    \hline\noalign{\smallskip}
  \end{tabular}
  \label{tab:megadepth}
\end{table}
\noindent\textbf{Results.}
We resize the images with the long side to $1200$ during the inference and follow the same evaluation protocol as indoor pose estimation.
As shown in Fig.~\ref{fig:comparision}, since the outdoor images contain less low texture regions and repetitive patterns, the unimodal model baseline (LoFTR) could also predict many correct correspondences for robust camera pose estimations.
However, the results in Table~\ref{tab:megadepth} indicate that our 3DG-STFM teacher model achieves better performance by leveraging the relative depth. 
The student model learned from the teacher could also outperform LoFTR, the state-of-art unimodal competitor. 
We find our student model averagely detects \textbf{1864.63} inlier (epipolar error less than $1\times10^{-4}$ in the normalized image coordinates) correspondences on each outdoor image pair, which is also higher than LoFTR's \textbf{1694.60} inlier detections.

\subsection{Homography Estimation} \label{sec:home_est}
We also evaluate our student model for homography estimation on HPatches dataset~\cite{hpatches}.
Following previous work~\cite{superglue,loftr}, we select 108 image sequences under large illumination changes or significant viewpoint variations for evaluation.
Every test image sequence contains one reference image and five pairing images.

\noindent\textbf{Evaluation Protocol.} 
We resize the original images with shorter dimensions equal to 480 and find the top 1K correspondences for each pair for detector free methods.
Our 3DG-STFM student model is trained on Megadepth~\cite{megadepth} mentioned in Section~\ref{sec:outdoor}.
All baseline results are reported using their original default implementation hyperparameters.
Homography estimation is performed by the OpenCV RANSAC implementation.
Following~\cite{superpoint}, we compute the reprojected mean error of the four corners of the image and report the area under the cumulative curve (AUC) up to three values: 3, 5, and 10 pixels in Table~\ref{tab:homography}.

\begin{table}[t]
  \centering
  \setlength{\tabcolsep}{6pt}
  \caption{\textbf{Homography estimation on HPatches~\cite{hpatches}.} The AUC of the corner error in percentage is reported.}
  \begin{tabular}{cccccc}
    \hline
    \multirow{2}{*}{Category} & \multirow{2}{*}{Method} & \multicolumn{3}{c}{Homography est. AUC} &\multirow{2}{*}{\#matches}\\\cline{3-5}
    \multirow{2}{*}{} & \multirow{2}{*}{} &@3px &@5px&@10px&\multirow{2}{*}{}\\
    \hline
    \multirow{4}{*}{Detector-based} 
    &D2Net~\cite{d2net}+NN &23.2&35.9&53.6&0.2k\\
    &R2D2~\cite{r2d2}+NN &50.6&63.9&76.8&0.5k\\
    &DISK~\cite{disk}+NN &52.3&64.9&78.9&1.1k\\
    &SP~\cite{superpoint}+SuperGlue~\cite{superglue} &53.9&68.3&\textbf{81.7}&0.6k\\
    \hline
    \multirow{4}{*}{Detector free}
    & Sparse-NCNet~\cite{efficient_NCN} &48.9&54.2&67.1&1.0k \\
    &DRC-Net~\cite{dual} &50.6&56.2&68.3&1.0k\\
    & LoFTR~\cite{loftr} &63.4&71.9&79.9&1.0k\\
    & 3DG-STFM &\textbf{64.7}&\textbf{73.1}&81.0&1.0k\\
    \hline
  \end{tabular}
  \label{tab:homography}
\end{table}

\noindent \textbf{Results.} 
Our 3DG-STFM student model is generalized well on the homography estimation task and achieves best performance compare with detector free methods as shown in Table~\ref{tab:homography}.
The method based on Superpoint~\cite{superpoint} and Superglue~\cite{superglue} get better performance at AUC$@10$px than our approach.
However, the 3DG-STFM student model shows more accurate performances under the other two strict metrics.
We provide more details in the supplementary material.

\subsection{Student-Teacher Learning Understanding}
\begin{figure*}[t]
  \centering
    \includegraphics[width=0.9\linewidth]{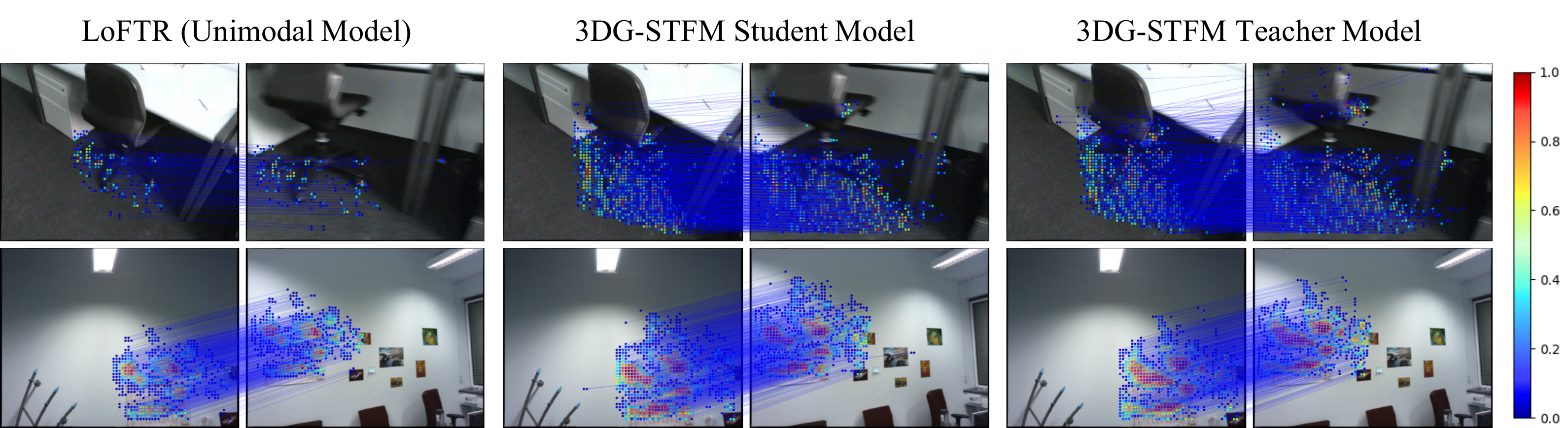}
  \caption{
  \textbf{Visualization of matching distribution change for better understand student-teacher learning.} The color of correspondence scatter is determined by the confidence score predictions of each model. The teacher model not only guide the student model to find more correspondences, but also teaches the confidence score distribution to the student model.}
  \label{fig:dist_trans}
\end{figure*}

\noindent\textbf{Visualizing Knowledge Transfer.}
To understand how our teacher model transfers knowledge to the student, we visualize the matching details to compare our student model and teacher model on ScanNet.
We remove the teacher branch and training student branch solely based on direct supervision and treat it as the vanilla unimodal model for comparisons. 
Since we adopt the LoFTR's matching strategy, this vanilla unimodal model is the same as LoFTR.
In Fig.~\ref{fig:dist_trans}, we plot all the predicted matches of models with a confidence score higher than $0.2$. We show in the first row of Fig.~\ref{fig:dist_trans} that both the teacher and student model find much more correspondences around low-texture regions than state-of-the-art. The teacher model explores the depth modality and then guides the student model to learn the RGB-induced depth information to increase the discriminant features in areas with low texture but depth variations. We also show that the student follows the teacher's confidence score pattern, while both have different patterns compared to LoFTR, shown in the second row of Fig.~\ref{fig:dist_trans}. The confidence scores are indicated by color, high in red, low in blue.
This knowledge transfer is achieved by proposed MQD loss and attentive loss for coarse-level and fine-level matching for our student-teacher architecture. 

\noindent \textbf{Ablation Study.}
To better understand the contribution in each module, we randomly select 150 scenes from ScanNet as a mini version dataset and test different variants of our model.
The test set is the same as the original ScanNet.
As shown in Table~\ref{tab:ablation}, the teacher model achieves the best performance and is used to teach the two models, i.e., Unimodal+MQD and Unimodal+MQD+Att.
The unimodal model is trained under direct supervision provided by dense correspondences labels based on Equation~\ref{eq:teacher}.
Compared with the unimodal model, both MQD and attentive loss help the knowledge transfer from teacher to student.
\begin{table}[t]
 \centering
 \setlength{\tabcolsep}{4pt}
  \begin{minipage}[c]{0.49\textwidth}
  \caption{\textbf{Ablation study.}}
  \resizebox{\linewidth}{!}{
  \begin{tabular}{cccc}
     \hline
     \multirow{2}{*}{Method} & \multicolumn{3}{c}{Pose estimation AUC}\\\cline{2-4}
     \multirow{2}{*}{} &$@5^{\circ}$ &$@10^{\circ}$ &$@20^{\circ}$\\
     \hline
     Multi-model Teacher &18.41~&36.53&54.07\\
     \hline
     Unimodal &14.78~&31.47&48.44\\
     Unimodal+MQD &16.46~&33.62&51.70\\
     Unimodal+MQD+Att &\textbf{17.05}~&\textbf{34.77}&\textbf{52.26}\\
    \hline
  \end{tabular}
  }
  \label{tab:ablation}
  \end{minipage}
  \hfill
  \begin{minipage}[c]{0.48\textwidth}
  \renewcommand{\arraystretch}{0.93}
  \caption{\textbf{Model compression study.}}
  \resizebox{\linewidth}{!}{
  \begin{tabular}{@{}cccccc@{}}
     \hline\noalign{\smallskip}
     \multirow{2}{*}{Method}
    &\multirow{2}{*}{$\mathrm{L}_{c}$}
     &\multirow{2}{*}{$\mathrm{L}_{f}$} & \multicolumn{3}{c}{Pose estimation AUC}\\\cline{4-6}
     \multirow{2}{*}{}
     &\multirow{2}{*}{}
     &\multirow{2}{*}{}
     &$@5^{\circ}$ &$@10^{\circ}$ &$@20^{\circ}$\\
     \hline\noalign{\smallskip}
     Teacher Model  &4&1&18.41&36.53&54.07\\
     Full-Size Student Model  &4&1&17.05&34.77&52.26\\
      \hline\noalign{\smallskip}
     Full-Size Model &4&1&\textbf{14.78}&31.47&48.44\\
     Slim Model &2&1&14.18&29.68&46.45\\
     Slim Student Model &2&1&14.49&\textbf{31.76}&\textbf{49.51}\\
     \hline\noalign{\smallskip}
  \end{tabular}
  }
  \label{tab:compression}
  \end{minipage}
\end{table}

\noindent\textbf{Model Compression Performance.}
Our architecture can also be generalized to model compression tasks.
We implement the model compression experiments on the mini version of ScanNet.
The results shown in Table~\ref{tab:compression} indicate our slim model has the competitive performance with the uncompressed model. 
In Table~\ref{tab:compression}, the slim matching model is proposed with half attention layers on the coarse-level transformer $\mathrm{L}_{c}$. The Slim Student Model is trained with our student-teacher architecture, while
both the Full-Size Model and the Slim Model are trained under the direct supervision provided by ground-truth. By learning knowledge from the Teacher Model, the Slim Student Model improves $2\%$ compared to the Slim Model and also shows better performance than the Full-Size Model at AUC$@10^{\circ}$ and $@20^{\circ}$.
More details are provided in the supplementary.

\section{Conclusion}
In this paper, we propose 3DG-STFM: a novel student-teacher learning framework for the dense local feature matching problem.
Our proposed framework mines depth knowledge from one multi-modal teacher model to guide the student model to learn the hidden depth information embedded in the RGB domain.
Two attentive mechanisms, i.e., MQD loss and attentive loss, are proposed to help the knowledge transfer.
Our student model is evaluated on several image matching and camera pose estimation tasks on indoor and outdoor datasets and achieves state-of-the-art performances.
Our 3DG-STFM also shows generalization ability on model compression tasks.

%
%
\bibliographystyle{splncs04}
\bibliography{egbib}
\end{document}


\pagestyle{headings}
\mainmatter
\def\ECCVSubNumber{7283}  

\title{Supplementary Materials} 

\titlerunning{3DG-STFM}
\author{Runyu Mao\inst{1}\and
Chen Bai\inst{2}\and
Yatong An\inst{2}\and
Fengqing Zhu\inst{1}\ and
Cheng Lu\inst{2}}
\authorrunning{R. Mao et al.}
\institute{Purdue University
\email{\{mao111, zhu0\}@purdue.edu}\and
XPeng Motors
\email{\{chenbai, yatongan, luc\}@xiaopeng.com}
}
\maketitle
\section{Homography Estimation}
We take a more detailed look at the homography evaluation from Section~4.3.
We follow~\cite{d2net} to select 108 image sequences.
There are 52 sequences under significant illumination changes (illumination set) and 56 sequences that have significant viewpoint variations (viewpoint set).
All images are resized with shorter dimensions equal to 480.
The model we used to evaluate is the student model we trained on Megadepth dataset~\cite{megadepth}, the training setting is reported in Section 3.5.
We select top $1000$ matches and vary the threshold from 1 pixel to 10 pixels for Mean Match Accuracy (MMA)~\cite{mma} calculation.
As shown in Table~\ref{tab:mma}, we evaluate our model on illumination set and vewpoint set seperately and finally report the overall performance on the entire 540 image pairs.
\begin{table}[ht]
  \centering
  \setlength{\tabcolsep}{3pt}
  \caption{\textbf{Evaluation on HPatches~\cite{hpatches}.} The Mean Match Accuracy (MMA) at different thresholds.}
  \begin{tabular}{ccccccccccc}
    \hline
    \multirow{2}{*}{Test set} &  \multicolumn{10}{c}{MMA (\%) threshold}\\\cline{2-11}
    \multirow{2}{*}{} &$\leq$1px &$\leq$2px&$\leq$3px&$\leq$4px&$\leq$5px&$\leq$6px&$\leq$7px &$\leq$8px&$\leq$9px&$\leq$10px\\
    \hline
    Illumination &79.50 &95.18 &98.58 &98.79 &98.87 &98.98 &99.14 &99.33 &99.48 &99.56\\
    \hline
    Viewpoint &55.78 &74.14 &79.01 &80.98 &81.85 &82.29 &82.54 &82.74 &82.90 &83.02\\
    \hline
    Overall &67.53 &84.57 &88.71 &89.81 &90.28 &90.56 &90.77 &90.96 &91.12 &91.22\\
    \hline
  \end{tabular}
  \label{tab:mma}
\end{table}

\section{Model Compression}
Our student-teacher learning architecture can also be generalized to model compression tasks.
A slim model with half attention layers on the coarse-level transformer $L_{c}$ is introduced.
In this section, we compare our full-size model and slim model on other metrics, i.e. FLOPs and total parameters, to see the compression effects.
The model contains Feature Pyramid Networks (FPN), coarse-level transformer, and fine-level transformer. The coarse-level threshold is removed so that all the coarse matching pairs will be fed to the fine-level transformer.
The input is a random tensor with $3\times640\times480$ dimension.

\begin{table}[h]
  \centering
  \setlength{\tabcolsep}{6pt}
  \caption{\textbf{Model Compression Performances.}}
  \begin{tabular}{ccccc}
    \hline
    Model &$L_{c}$&$L_{f}$&FLOPs(B) & Params (M) \\
    \hline
    Full-size model &4&1&365.23&11.57 \\
    \hline
    Slim model &2&1&325.71&8.95 \\
    \hline
  \end{tabular}
  \label{tab:flops_parameter}
\end{table}
As shown in  Table~\ref{tab:flops_parameter}, we reduce the coarse-level transformer layer by half for our slim model. The overall FLOPs are reduced by $10.82\%$ and the total parameters are reduced by $22.64\%$.
\section{Qualitative Results}\label{results}
\begin{figure}
  \centering
   \includegraphics[width=0.83\linewidth]{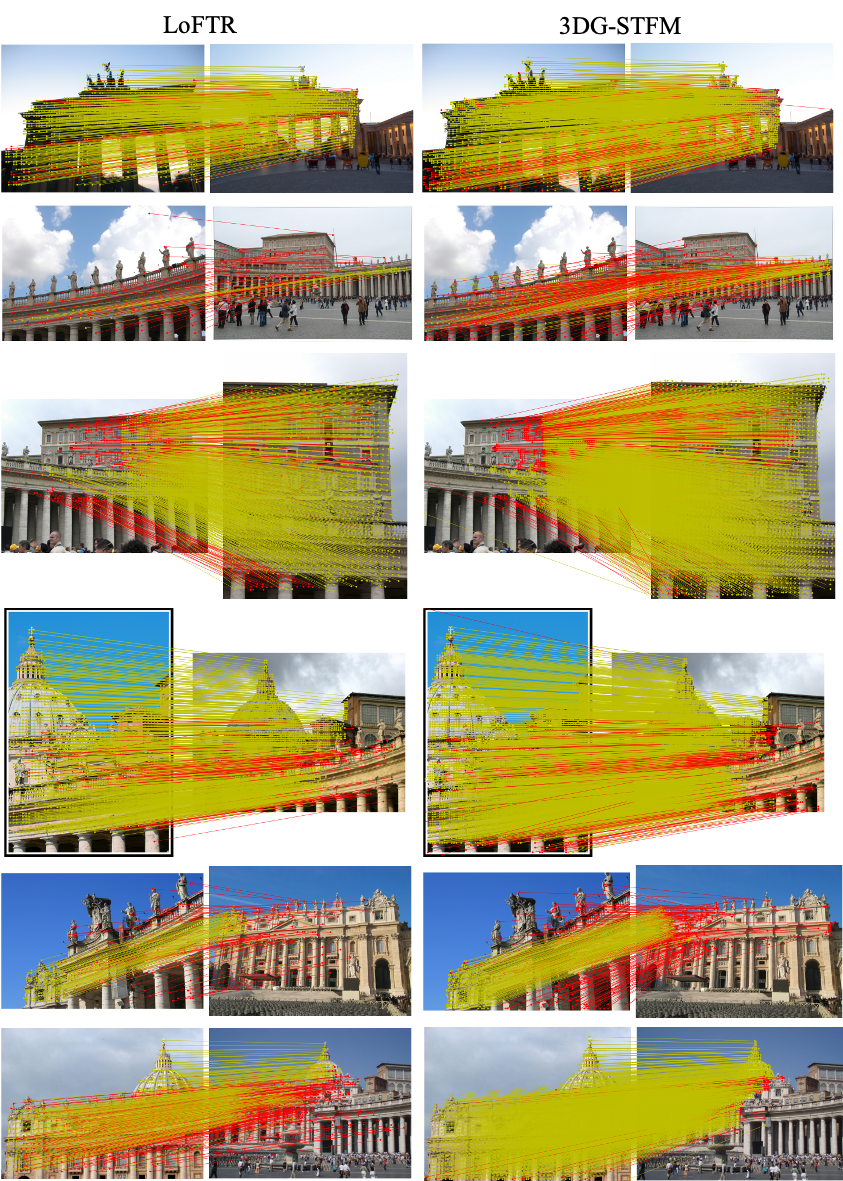}

   \caption{\textbf{Qualitative image matches on Megadepth~\cite{megadepth}.}The red color indicates epipolar error beyond $1\times 10^{-4}$ (in the normalized image coordinates).}
   \label{fig:megdepth}

\end{figure}

More comparison of our 3DG-STFM and baseline
LoFTR~\cite{loftr} on ScanNet~\cite{scannet} and MegaDepth~\cite{megadepth} datasets is illustrated
in Fig.~\ref{fig:scannet} and~\ref{fig:megdepth} We also include a video to demonstrate
our method’s performance on a challenging low texture
indoor scene. The scatter indicates the feature location
and the color represents the confidence score, high
in red, low in blue. Benefited from the discrimination of
depth modality, our method could detect more features on
low texture regions since it learns the RGB-induced depth
distribution from the teacher model.
\begin{figure*}
  \centering
  \vspace{-15pt}
      \includegraphics[width=0.86\linewidth]{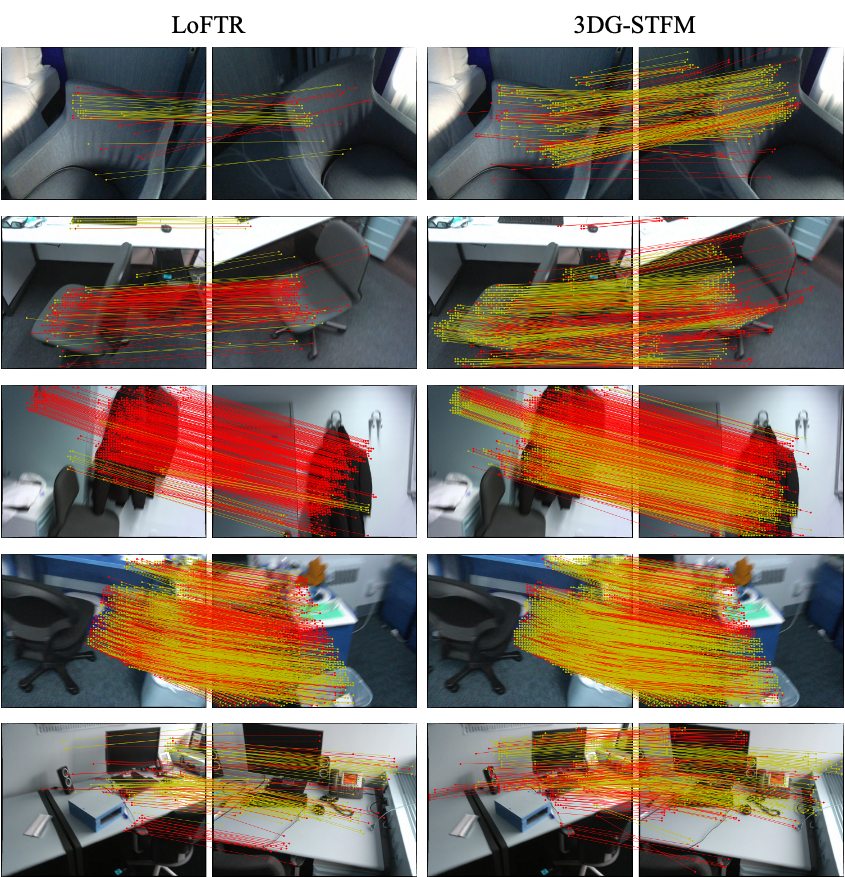}
   \caption*{}
   \label{fig:scannet}

\end{figure*}
\vspace{-3pt}
\begin{figure}
  \centering
      \includegraphics[width=0.86\linewidth]{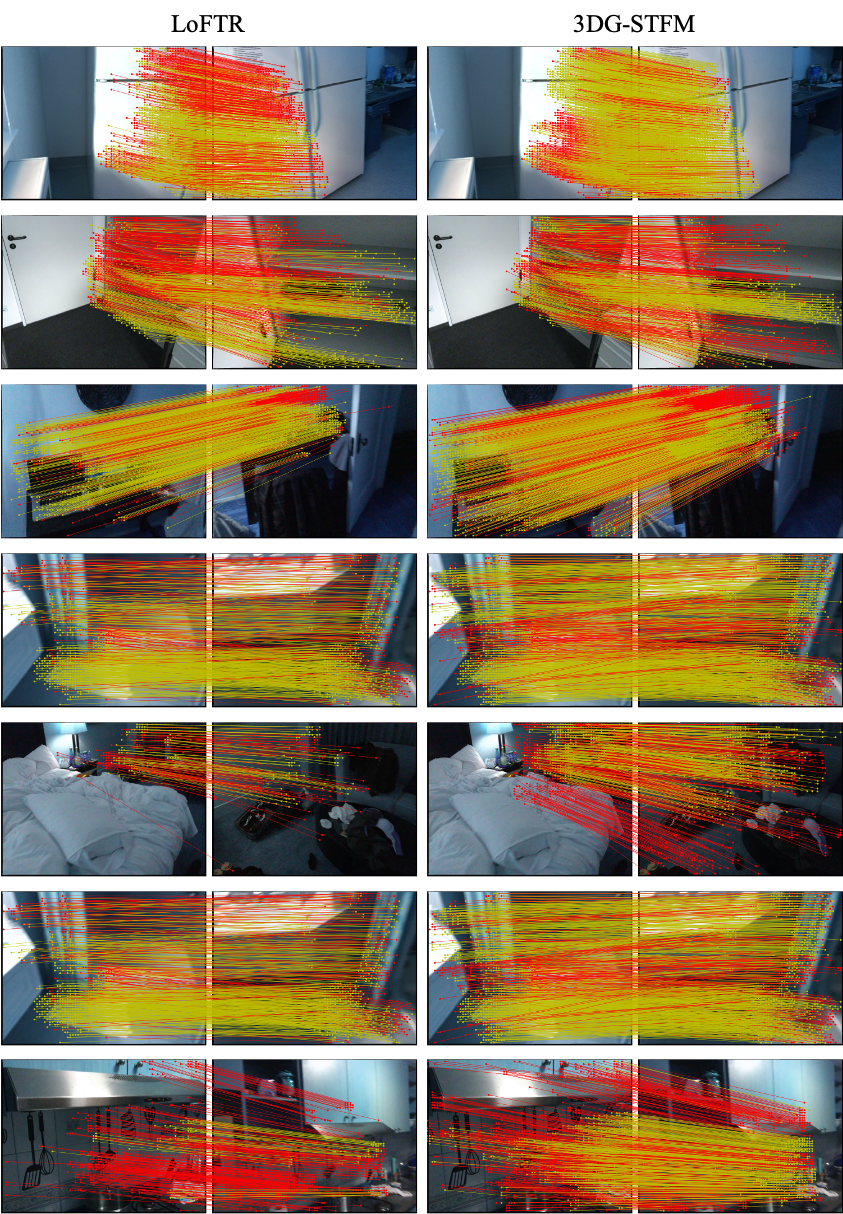}
   \caption{\textbf{Qualitative image matches on ScanNet~\cite{scannet}.} The red color indicates epipolar error beyond $5\times10^{-4}$ (in the normalized image coordinates).}
   \label{fig:scannet}

\end{figure}

\begin{figure*}
  \centering
  \vspace{-15pt}
      \includegraphics[width=0.86\linewidth]{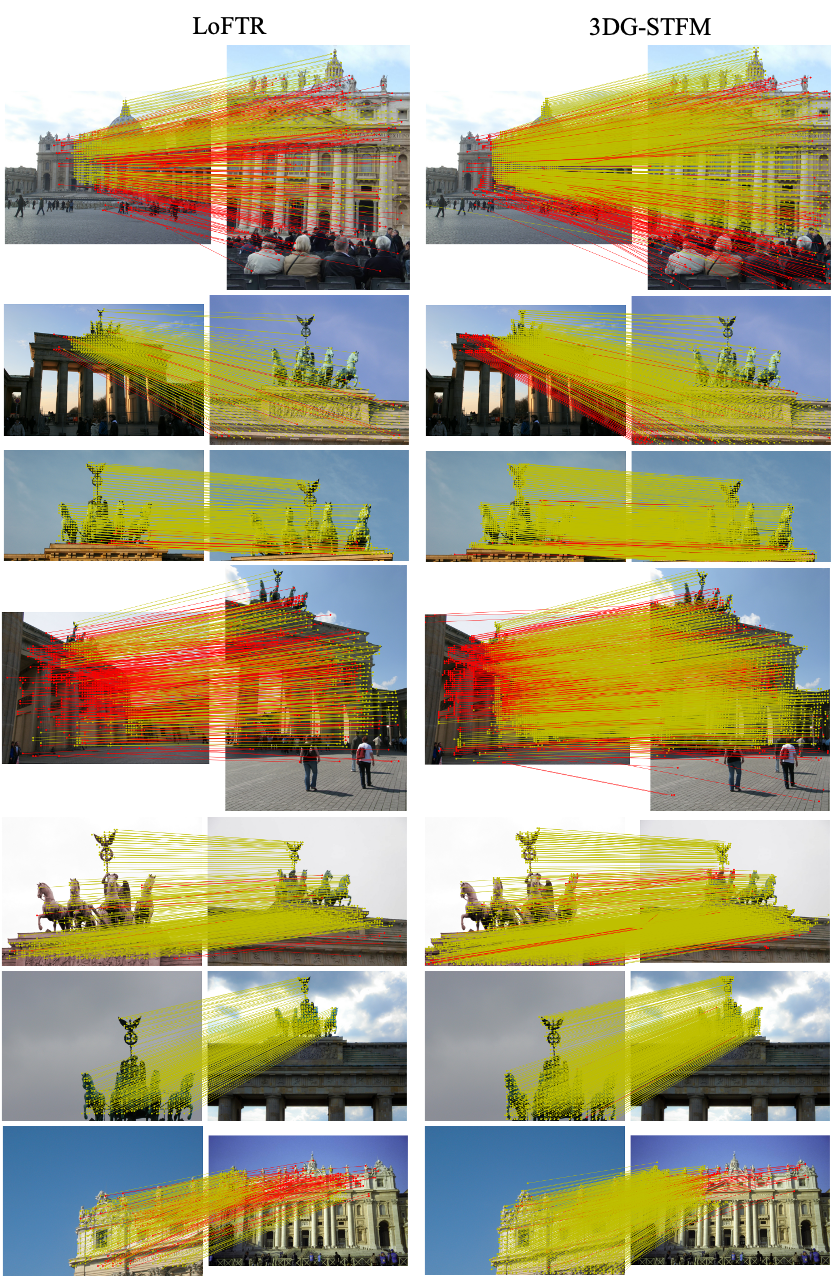}
   \caption*{}
   \label{fig:megdepth_p1}

\end{figure*}

{\small
\bibliographystyle{splncs04}
\bibliography{egbib}
}